# Gaze estimation problem tackled through synthetic images

Gonzalo Garde
Public University of Navarre
Pamplona, Spain
gonzalo.garde@unavarra.es

Andoni Larumbe-Bergera
Public University of Navarre
Pamplona, Spain
andoni.larumbe@unavarra.es

Benoît Bossavit
Trinity College Dublin
Dublin, Ireland
bossavib@scss.tcd.ie

Rafael Cabeza
Public University of Navarre
Pamplona, Spain
rcabeza@unavarra.es

Sonia Porta
Public University of Navarre
Pamplona, Spain
sporta@unavarra.es

Arantxa Villanueva
Public University of Navarre
Pamplona, Spain
avilla@unavarra.es

## ABSTRACT

In this paper, we evaluate a synthetic framework to be used in the field of gaze estimation employing deep learning techniques. The lack of sufficient annotated data could be overcome by the utilization of a synthetic evaluation framework as far as it resembles the behavior of a real scenario. In this work, we use U2Eyes synthetic environment employing I2Head datataset as real benchmark for comparison based on alternative training and testing strategies. The results obtained show comparable average behavior between both frameworks although significantly more robust and stable performance is retrieved by the synthetic images. Additionally, the potential of synthetically pretrained models in order to be applied in user's specific calibration strategies is shown with outstanding performances.

## 1 INTRODUCTION

Deep learning solutions have produced a revolution in the field of computer vision. These learning models are gaining momentum in eye tracking and gaze estimation as clearly demonstrated in the bunch of papers published in the last years [He et al. 2019] [Linden et al. 2019] [Guo et al. 2019]. One of the most important requirements of deep learning models is the availability of large scale annotated datasets. However, annotated data are sometimes comparatively insufficient, which makes it an obstacle to achieve generalization requirement. This can be applied to those models employed for eye landmarks detection and to those devoted to estimating gaze.

Obtaining annotated data in eye tracking is not trivial. Annotation based on manual labelling is far from being a practical solution. In many computer vision problems, using synthetic images as data augmentation technique has demonstrated to be a useful solution to enable researchers to cope with the problem of insufficient data [Lee and Moloney 2017] [Lindner et al. 2019]. Apart from the availability of huge amount of data, the possibility of providing labels at the same time is one of the main advantages of the simulation frameworks. Similarly, in the field of eye tracking several efforts have been oriented to generate synthetic training data [Wood et al. 2016] [Świrski and Dodgson 2014] [Porta et al. 2019].

The degree of resemblance between synthetic and real data is argued in this paper. Synthetic data and original data should be statistically similar in order to ensure a fair comparison. Additionally, machine learning techniques employing real data should present comparable results when using synthetic data so that methods can be directly reused on synthetic data, and vice versa. If these requirements are fulfilled, the synthetic framework is validated and the potential of the simulated environment becomes significantly important in the field. Transfer learning would be an immediate consequence of such a result. One of the problems of deep learning using CNN is that the learning stage, during which weights are learnt, can be very time-consuming, moreover, a considerable number of images is needed. Transfer learning permits to reuse a pre trained network to be applied in a new problem for which the original network was not trained. Transfer learning techniques are varied but basically the idea behind would be to use pre trained model and to perform a simpler training process using fewer data representing the new problem. The assumption is somehow grounded by the existence of common features (blobs, edges...) in the images from natural scenes that are common to different problems. It can be expected a higher success of the transfer learning procedure as the similarity between the datasets increases.

In this paper we employ two datasets to validate the framework for gaze estimation purposes: first a synthetic binocular images dataset, named as U2Eyes [Porta et al. 2019], is employed, second, the real data I2Head [Martinikorena et al. 2018] dataset is used as comparison benchmark. A network based on ResNet-18 is proposed as testing scenario [He et al. 2015].

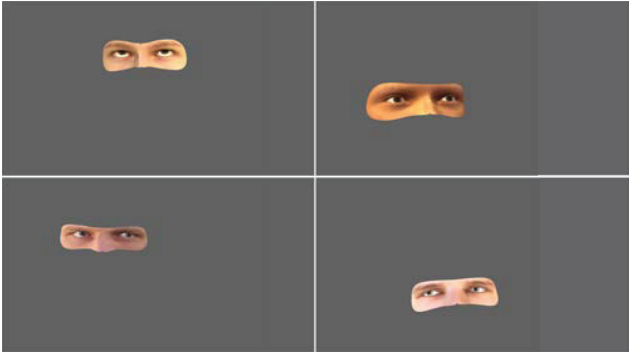

Figure 1: Samples extracted from U2Eyes dataset.

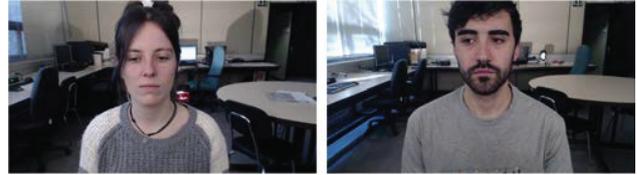

Figure 2: Samples extracted from I2Head dataset.

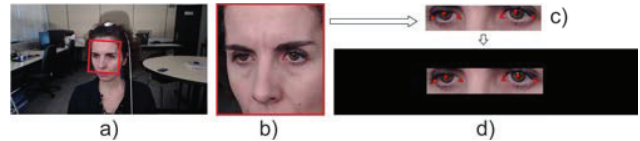

Figure 3: Image preprocessing. The images (a) are rotated (b) so eyes are in the same line, which makes the bounding box (c) contains as meaningful information as possible. Then, a padding process (d) assures that all images have the same size before feeding them to the network.

We are not pursuing the best CNN-based model for gaze estimation but to validate a synthetic framework for gaze estimation and future research works. The contributions of this paper are:

- To show the validity of using synthetic data as a mean to train networks employing CNN that will be applied to real data in the gaze estimation field.
- To highlight the potential applications of transfer learning in gaze estimation where the learning stage is based on simulated data that can be applied to real inputs.

The paper is organized as follows, in section 2 both, the synthetic and real datasets are described. In section 3 the architecture proposed is shown. The validation experiments are carried out in section 4. Finally, the conclusions of the work are presented.

## 2 DATASETS

The proposed synthetic framework is based on U2Eyes dataset [Porta et al. 2019]. Twenty different simulated subjects are provided in the public version of the dataset. The images have a resolution of 3840x2160 pixels (4K) and were created using Unity. The provided images represent eyes area of a subject gazing at different points on a screen simulating a standard remote eye tracking session. The images are annotated with head pose and observed points information. Additionally, 2D and 3D landmarks information of both eyes is provided. For each subject 125 head positions are simulated from which two gazing grids containing 15 and 32 points are observed. Consequently, 5,875 images per user are provided resulting in a total of approximately 120K images. U2Eyes represents a rich appearance variation environment. The authors claim that essential eyeball physiology elements and binocular vision dynamics have been modelled. In figure 1 samples of the dataset are shown.

I2Head dataset is employed as a real benchmark in order to validate the proposed framework [Martinikorena et al. 2018]. I2Head is a public dataset providing images annotated with gaze and head pose data. The dataset contains information about twelve individuals gazing two grids containing 17 and 65 points from eight different head positions, in constrained and free head movement scenarios. I2Head lacks from image landmark information. In figure 2 samples of the dataset are shown. For each user, information about 232 ($65 \times 2 + 17 \times 6$) gazing points is provided representing a total of 2,784 points data.

In order to perform a realistic comparison between the datasets, eye area is extracted from I2Head images. Since 2D data are not provided with the dataset, a manual labelling of the eye region bounding box has been performed for a reduced set of images of the dataset. More specifically, the four central sessions have been selected for each user containing 164 gazing points per user.

U2Eyes resembles to a large extent I2Head dataset. Both databases represent standard eye tracking sessions in a remote setup. However some differences are found. While the camera is positioned on the top of the screen in I2Head dataset a centered position with respect to the gazing grid is selected in U2Eyes. Consequently, in U2Eyes subjects gaze points above the camera while all the points are placed below the camera in the I2Head dataset. In the same manner, the range of head poses is different in both datasets when referring to vertical positions. All these aspects should be taken into account when the alternative training strategies are proposed.

## 3 GAZECNN

### 3.1 Data preprocessing

The preprocessing applied to U2Eyes and I2Head data is described: both synthetical and real image preprocessing is equivalent, and it is shown in figure 3. First, the original image, (a), is rotated (b) in order to normalize the roll component. The rotation is done using the angle between the horizontal and the straight line defined by the two outer eye corners. Then, a bounding box is created using the distance between the two outer eye corners and the image is cropped (c). The rotation before cropping is necessary to prevent the gray pixels that would be present in the U2Eyes images if we would just obtain the bounding box from the original source. To maintain the similarity for both datasets, the same procedure is applied to I2Head images. Finally, a black edge is added until the image size is 390x85 pixels (d). As the images from both datasets cover a range of well-known distances, the black edge was added to make all images the same size. Thus, the farther to the camera,

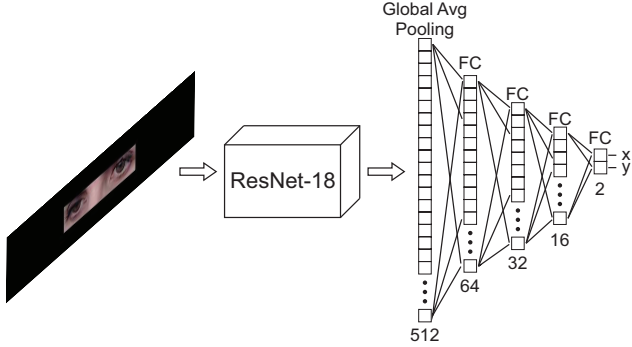

Figure 4: Architecture proposed. The backbone of the network consist on a Resnet-18 to extract meaningful characteristics from the image. Then, these characteristics are fed into fully connected regressor network to obtain the final gaze components.

the bigger the black edge, providing additional depth information to the network.

### 3.2 Architecture description

The architecture of our network consists on the Resnet-18 [He et al. 2015] as the base of the model, followed by a Global Average Pooling layer and three Fully Connected Layers. The choice of the Resnet-18 as core of our architecture is supported by its performance over Imagenet [Deng et al. 2009] classification task while being simple enough to make retraining steps feasible in both, time and hardware requirements. The results obtained over Imagenet ensure that the network is able to extract meaningful features from images. The three fully connected layers at the top of the network make use of these features to compute the x and y coordinates of the look at points.

## 4 EXPERIMENTS

In this section, the potential application of the virtual environment to real images in gaze estimation is evaluated. Gaze estimation is a regression problem and our objective is to measure the performance and robustness of our network when the subject fixates a point. To this end, the angular offset between the estimated gaze direction and subject's visual axis is used as evaluation metric. This angular offset is calculated by computing the distance between the estimated point and the real point in the grid and the known distance between the real point and the camera. For each trained model, the error among estimated and ground truth points is typified by its statistical distribution. To test the validity of the synthetic dataset as a suitable workspace for gaze estimation problems, different experiments were designed. The differences among experiments rely on the dataset used to train the final model and the initial weights of the network, Imagenet or the combination of Imagenet and U2Eyes in our case. The experiments carried out are summarized in table 1.

To assure the independence of the obtained results, Leave-One-Out strategy was used for the different trainings, i.e. in cases 2 and 3 the experiment was run as many times as number of users in I2Head dataset, leaving one of the users out for testing in every run and training with the others, and averaging the results to obtain the final outcome. The aim pursued in cases 4 and 5 was to simulate the standard calibration in which the model is adapted to a unique subject. Calibration is a well-known procedure that is understood as the process to construct a person-specific gaze tracker during which particular characteristics of the individual are learnt by the system. In this paper, we conduct subjects' calibration as a meta-learning procedure. The pre trained network is fine-tuned employing user's specific data in a fewer shot adaptive gaze estimation approach [He et al. 2019]. The weights of the network are modified using two sessions of each individual (calibration) while the other two are employed for testing purposes.

| Case | Initial weights | Train | Test |
|---|---|---|---|
| 1 | Imagenet | U2Eyes (19 users) | U2Eyes (1 user) |
| 2 | Imagenet+U2Eyes | I2Head (11 users) | I2Head (1 user) |
| 3 | Imagenet | I2Head (11 users) | I2Head (1 user) |
| 4 | Imagenet+U2Eyes | I2Head (1 user) | I2Head (1 user) |
| 5 | Imagenet | I2Head (1 user) | I2Head (1 user) |

Table 1: Methods comparison. The initial weights column refers to the weights the model had before the training started. In the case of Imagenet + U2Eyes, it refers to the weights after the synthetic training (Case 1). The train and test columns focus on the train and test sets distributions in every run of the Leave-One-Out strategy for U2Eyes (Case 1) and I2Head (Case 2, 3, 4, 5).

| Case | Mean(°) | Std(°) | Median(°) |
|---|---|---|---|
| 1 | 1.94 | 1.21 | 1.92 |
| 2 | 2.73 | 1.98 | 2.26 |
| 3 | 2.59 | 1.91 | 2.18 |
| 4 | 3.54 | 2.47 | 3.00 |
| 5 | 15.11 | 7.83 | 15.37 |

Table 2: Methods comparison in gaze angular error. Special attention to the similarity among cases 2 and 3, and the gap between cases 4 and 5.

Gaze angular error is provided as performance measurement of the experiments that are summarized in table 2.

Additionally error distribution can be more easily compared in figure 5. The results for cases 2 and 3 are comparable, thus meaning that introducing U2Eyes does not yield significant differences. However, the results obtained in case 1 in which U2Eyes is solely used permits us to conclude that a more compact distribution is obtained free of outliers. Although the synthetic environments can approach a real scenario realistically, they provide a more controlled and stable framework. Even if comparable average values can be observed, the values obtained in figure 5 show that more robust results can be expected in the synthetic setup compared to cases 2 and 3. On the other hand, an outlier filtering process applied to cases 2 and 3 would rely in a more comparable behavior of these cases. We would like to make clear that no hypertuning has been carried out in any of the cases.

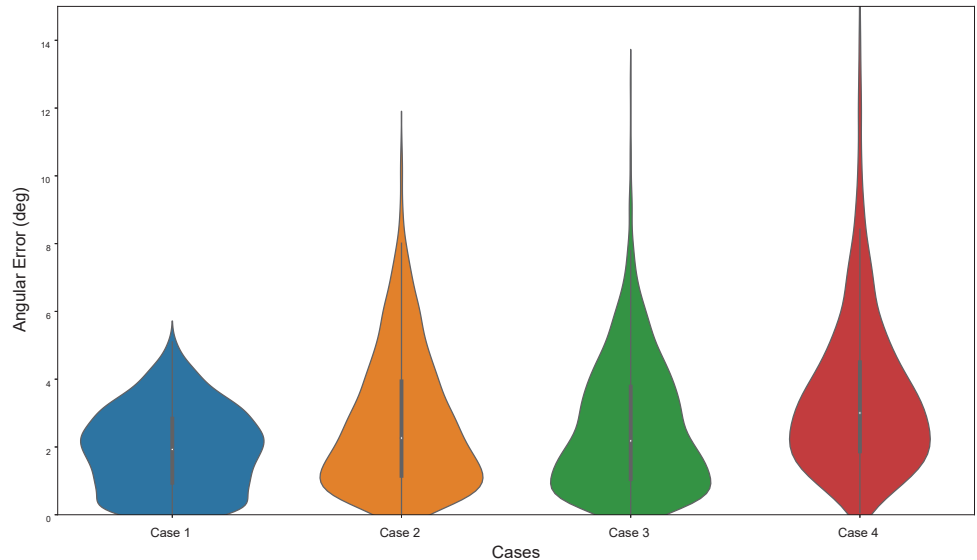

**Figure 5: Performance comparison of cases 1 to 4. The horizontal axis corresponds to the cases. The vertical axis shows the angular error in degrees. The figure represents the distribution of each case in a violin plot.**

On the other hand, one of the most important conclusions is the potential use of the synthetically pre trained models in order to obtain personalized calibrated models using a reduced set of individuals' images. Cases 4 and 5 provide an insight of the possibility of enhancing the performance of deep learning models for user calibration. A clear demonstration of the problems derived from the lack of enough data can be concluded from case 5. The number of images employed does not allow the model to obtain an acceptable performance confirming the necessity of pursuing solutions to increase the amount of annotated data in deep learning gaze estimation. However, when pre-training the network in a synthetic environment (case 4), the improvement is significant and closer to SotA (State of the Art).

## 5 CONCLUSIONS

Gaze estimation using deep learning techniques is an open topic today, being the lack of sufficient data one of the obstacles to overcome. In this paper, the validity of a synthetic framework for testing deep learning techniques is evaluated. The results show that the synthetic environment resembles to some extent the results obtained for a real scenario. Although average values are comparable, the robustness of the synthetic setup provides more compact distributions. On the other hand, the paper shows potential applications of synthetic data in order to pre train deep learning models that can be fine-tuned for a specific user emulating the well-known calibration procedure by using a reduced set of individual's real images. Finally, we can conclude that networks using Imagenet weights, trained with a huge number of real images, represent an interesting start point for gaze estimation models training.

## ACKNOWLEDGMENTS

We would like to acknowledge the Spanish Ministry of Science, Innovation and Universities for their support under Contract TIN2017-84388-R.

We gratefully acknowledge the support of NVIDIA Corporation with the donation of the Titan X Pascal GPU used for this research.

We thank Dr. Ahmed Abass (University of Liverpool) for his valuable data and answers.